\documentclass[10pt,twocolumn,letterpaper]{article}

\usepackage{iccv}
\usepackage{times}
\usepackage{epsfig}
\usepackage{graphicx}
\usepackage{amsmath}
\usepackage{amssymb}

\usepackage{xcolor} % for color
\usepackage{pifont}
\newcommand{\xmark}{\ding{55}}%
\newcommand{\graycross}{\textcolor{red}{\xmark}}
\newcommand{\greencheck}{\textcolor{blue}{\checkmark}}
\usepackage{booktabs}
\usepackage{float}
\usepackage{caption}
\usepackage{subcaption}
\usepackage{etoolbox}
\usepackage{multirow}
\AtEndEnvironment{table}{\vspace{-0.75cm}}
\AtEndEnvironment{figure}{\vspace{-0.5cm}} 
\AtEndEnvironment{figure*}{\vspace{-0.5cm}} 
\usepackage{colortbl}
\usepackage[font=small,skip=0pt]{caption}
\usepackage{comment}
\usepackage[pagebackref=true,breaklinks=true,letterpaper=true,colorlinks,bookmarks=false]{hyperref}

\iccvfinalcopy % *** Uncomment this line for the final submission

\def\iccvPaperID{4} % *** Enter the ICCV Paper ID here

\ificcvfinal\fi
\begin{document}
\captionsetup[subfigure]{skip=0pt}
%%%%%%%%% TITLE
\title{\textcolor{magenta}{UPGPT}: \textcolor{magenta}{U}niversal Diffusion Model for \textcolor{magenta}{P}erson Image \textcolor{magenta}{G}eneration, Editing and \textcolor{magenta}{P}ose \textcolor{magenta}{T}ransfer} 
%\title{UPGPT:  Universal Diffusion Model for Person Image Generation, \\ Editing and Pose Transfer}

\author{Soon Yau Cheong\\
University Of Surrey\\
{\tt\small s.cheong@surrey.ac.uk}
% For a paper whose authors are all at the same institution,
% omit the following lines up until the closing ``}''.
% Additional authors and addresses can be added with ``\and'',
% just like the second author.
% To save space, use either the email address or home page, not both
\and
Armin Mustafa\\
University of Surrey\\
{\tt\small armin.mustafa@surrey.ac.uk}
\and
Andrew Gilbert\\
University of Surrey\\
{\tt\small a.gilbert@surrey.ac.uk}
}

\maketitle

% Remove page # from the first page of camera-ready.
%\ificcvfinal\thispagestyle{empty}\fi

%%%%%%%%% ABSTRACT
\begin{abstract}
Text-to-image models (T2I) such as StableDiffusion have been used to generate high quality images of people. However, due to the random nature of the generation process, the person has a different appearance e.g. pose, face, and clothing, despite using the same text prompt. The appearance inconsistency makes T2I unsuitable for pose transfer. We address this by proposing a multimodal diffusion model that accepts text, pose, and visual prompting. Our model is the first unified method to perform all person image tasks - generation, pose transfer, and mask-less edit. We also pioneer using small dimensional 3D body model parameters directly to demonstrate new capability - simultaneous pose and camera view interpolation while maintaining the person's appearance. 

\end{abstract}
%\let\thefootnote\relax\footnotetext{Code and demo available at github.com/soon-yau/upgpt}
%%%%%%%%% BODY TEXT
\section{Introduction}

Generating humans from text and/or pose is a challenging problem in computer vision. The methods can be classified into two categories, (1) image generation (synthesis) and (2) pose transfer and editing. Image generation can be unconditional or conditioned on other information \eg, pose and text. Pose-guided image generation, conditions on pose (keypoints, skeleton image, heatmap, body mesh) to generate images \cite{pix2pix, vqgan, gaugan, AlBahar2021}; and text-to-image models such as DALL-E\cite{dalle,dalle2} and \cite{stackgan,attngan, dmgan, dfgan, xmcgan, imagen}. Pose or text to image is a one-to-many mapping. It can create a person with vastly different appearances even given the same conditions - \eg, a person in the same pose but wearing other clothing or shades of color from the word ``red shirt.'' The ambiguity and inconsistency prohibit them from being used to perform image editing, which requires the maintenance of the visual appearance of all other aspects of the image apart from the elements or regions to be edited. Some newer methods \cite{kpe, text2human, human_diffusion} use pose and text to exert further control. However, the effect is still limited by the inherent ambiguity of these modalities and is hence unsuitable for image editing.

\begin{figure}[!tb]
    \begin{center}
        \includegraphics[width=0.8\linewidth]{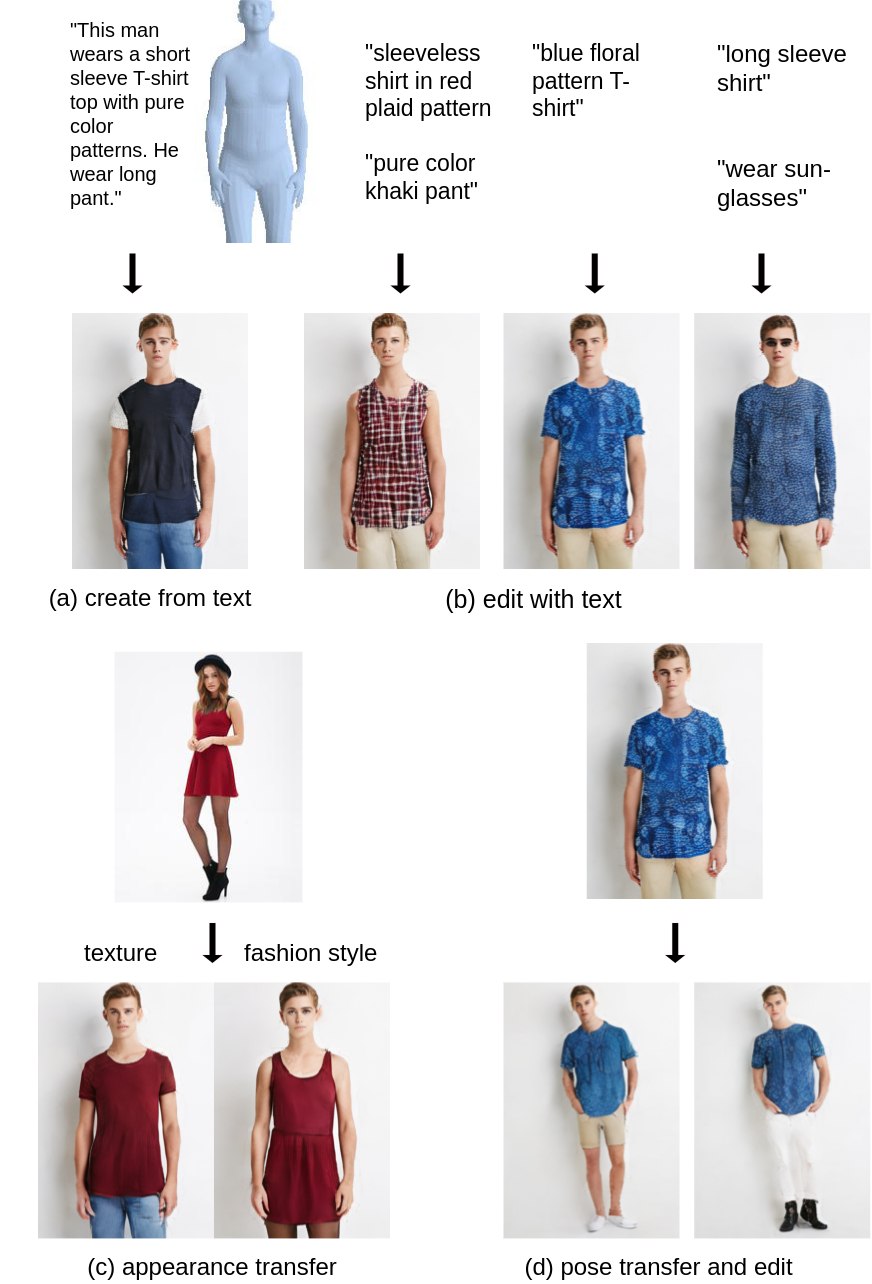}
    \end{center}
\caption{UPGPT can perform all person image generative tasks: (a) text and pose guided image generation, (b) fine-grained, mask-less region editing with text, (c) style and appearance transfer, (d) pose transfer followed by edit.} 
\label{fig:banner}
\vspace{-1mm}
\end{figure}
The other category is image editing, for tasks such as changing the clothing, human pose, or face. Most pose-guided image generation literature fall into this category, performing pose transfer to transfer a person's appearance from a source image to the pose of a target image. However, we prefer the term \emph{edit} to encompass other forms of modification, including using text or modifying the pose parameters directly, rather than having to \emph{transfer} them from the other image. Pose transfer models \cite{Ma2017,Siarohin2018,Yang2020} use both human pose and a source image as conditions for the generative image model where visual information of a source image serves as a stable condition to encourage the models to maintain a person's appearance in the generated image. \cite{patn,pise,nted,casd} have extended capabilities that could also transfer texture, clothing shape, or both, \ie, appearance transfer, but no single model can perform all those tasks. More importantly, they all need to train on source-target image pairs and can not generate a new image without a source image. To bridge the gap between the two categories, we propose \emph{UPGPT} to perform both generation and edit tasks using a single trained universal model, and image sampling pipeline, as seen in Figure \ref{fig:banner}.

\begin{figure}[t]
    \begin{center}
        \includegraphics[width=1.0\linewidth]{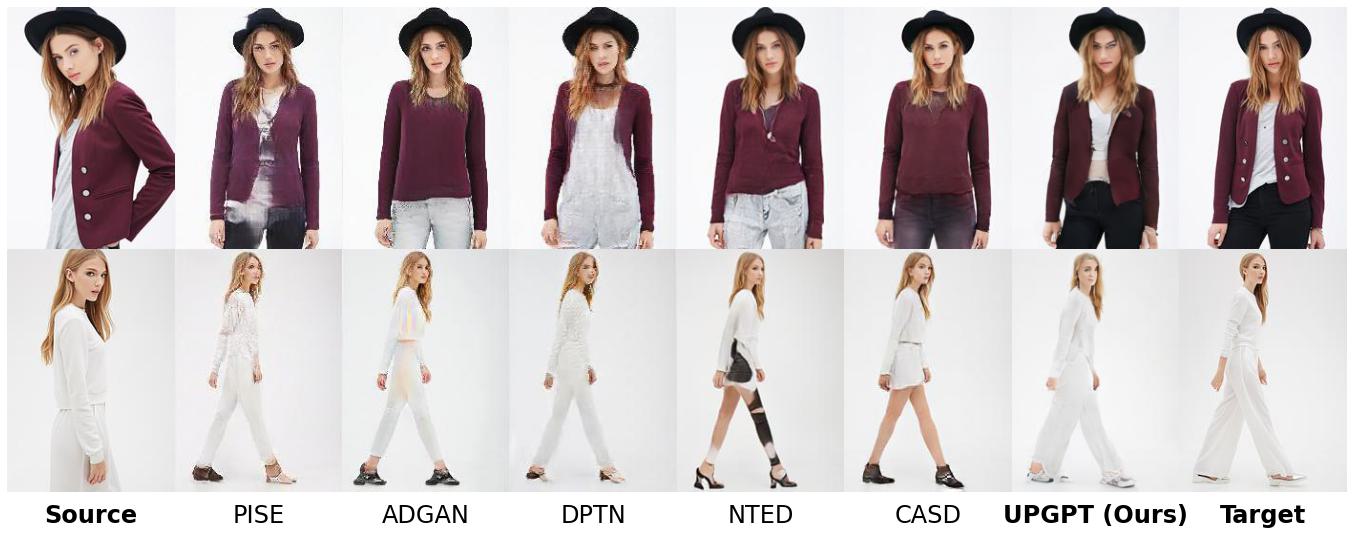}
    \end{center}
\caption{Pose transfer is an ill-posed problem: Often, the source image does not contain all information for the target pose \ie, in this figure, the pant. Compared to existing methods (PISE\cite{pise}, ADGAN\cite{adgan}, DPTN\cite{dptn}, NTED\cite{nted}, CASD\cite{casd}), our method can create the desired result by utilizing additional multimodal information.}
\label{fig:pt_missing}
\end{figure}

In our research, we discovered four underlying problems in person image generation and editing that have yet to be addressed: \textbf{(1)} Existing methods cannot interpolate human pose due to the inherent limitation of the chosen pose representations. 2D body segmentation map (parsing map) and body mesh are dense representations (pixel and voxel) and cannot be interpolated. To interpolate 2D keypoint points and their derivatives (skeleton image, heatmap), they must first be mapped into 3D space, which is a difficult task on its own, before performing the interpolation in 3D space, then project back into 2D keypoints. We break away from the tradition by using pose parameters of  SMPL\cite{smpl}, a 3D body model that represents pose by rotation of body joints. Then, performing linear interpolation on the SMPL parameters produces pose interpolation using our model. \textbf{(2)} Existing person image editing methods require parsing maps, which is difficult for users to create or edit by hand. Furthermore, their methods are typically constrained to transferring information from a single modality. To address this challenge, our method allows text or drag-and-drop of the reference image or a combination of them to perform convenient and fast image editing. %The other two problems we found related to the pose transfer task. The current de facto pose transfer task in computer vision literature follows \cite{patn}'s method to derive source-target image pairs from the DeepFashion dataset\cite{deepfashion}. Many image pairs are ill-posed and contain issues overlooked by existing literature: 
\textbf{(3)} Missing information from the source image. For example, when a target image expects a full-body person, the source image only contains a partial view where the lower part is not visible, as shown in Figure \ref{fig:pt_missing}. This leaves a question of whether the model should generate short pants, long pants, a dress, a sneaker, high heels, or leather shoes. \textbf{(4)} A person's appearance can change in the target image \eg, a person wearing a jacket in the source image may have it taken off in the target. Existing methods rely solely on the source image to provide all the information. Still, they can fail to generate desired or correct results if the information is incomplete or wrong, as shown in Figure \ref{fig:pose_transfer}. We address problems 3 and 4 by adding a new modality - text to enrich the information source and to reduce and correct errors. The text description of the expected outcome can work as a way to filter out unwanted information (not wearing a jacket) or to fill in missing information (to generate pants or skirts).

Table \ref{table:capabilities} compares the capabilities of the two main person image generation methods, and our proposed method combines all the key features. In summary, the main contributions of our papers are:
\begin{enumerate}
    \itemsep0em 
    \item A unified framework that can simultaneously perform person image generation, editing, and pose transfer tasks.
    \item The provision of zero-shot, mask-less image generation and editing with text. 
    \item The use of 3D parametric body model parameters to demonstrate the first simultaneous pose and camera view interpolation.
\end{enumerate}
\vspace{-3mm}
\begin{table}[!htb]
\vspace{-3mm}
\begin{center}
\begin{tabular}{l|c|c|c}
\toprule
& Pose  & Text-Pose-to & UPGPT  \\
& Transfer & Person Image & (Ours)\\
\toprule
Pose Edit & \greencheck{} & \graycross  & \greencheck \\
Appearance Edit & \greencheck & \graycross  & \greencheck \\
Texture Edit & \greencheck & \graycross & \greencheck \\
Create from Text & \graycross& \greencheck & \greencheck \\
Edit with Text & \graycross & \greencheck & \greencheck \\
Pose Interpolation & \graycross & \graycross & \greencheck \\
\bottomrule
\end{tabular}

\end{center}
\caption{Comparing the superset capabilities of pose transfer\cite{patn, Yang2020, gfla, adgan, pise, casd, nted}, text-pose-to-image \cite{kpe, human_diffusion, text2human} and our method. Our unified method can perform all the person generation and edit tasks and introduce a new capability of pose interpolation.}
\label{table:capabilities}
\end{table}

%-------------------------------------------------------------------------
\section{Related Works}
\textbf{Diffusion Models} (DM) \cite{diffusion_model, Dhariwal2021} have shown superior image quality and text-guided capability. In training, the DM gradually adds noise to the image until it becomes random noise; this process is known as forward diffusion. The diffused random noise act as latent variables and is denoised progressively to generate an image in image sampling; this progress is known as reverse diffusion. Typically, a UNet\cite{unet} is used to learn to produce the denoising signal. Most methods\cite{dalle2,imagen,glide} use the classified-free approaches\cite{Ho2021} to find the direction between the conditional and unconditional in the latent space, which is to be applied in sampling time to guide the model towards the conditioning direction. However, denoising every image pixel can be computationally expensive; therefore, LDM\cite{ldm} proposed a two-stage process. It first trained a variational autoencoder (VAE) \cite{vae} to encode the image into smaller dimensional latent variables, and the DM learned to produce the VAE latent variables. DMs could provide image editing by performing text-guided diffusion on regions defined by segmentation mask \cite{glide,dalle2, imagen,blended_diffusion,blended_latent_diffusion}. Dreambooth\cite{dreambooth} shows that they could encode a person's face into a text token and use a DM to generate the person in a different scene. More recently, \cite{prompt2prompt} proposed a mask-less edit of coarse objects by learning the region from the attention map. Our method achieves mask-less editing by learning and disentangling a person's appearance.

\textbf{Pose Guided Image Generation.}
Ma \etal \cite{Ma2017} was among the first literature on pose transfer; they concatenated source images with the target pose heatmap and used them as input conditions to a GAN\cite{gan}. Starting from PATN\cite{patn}, models take pose from both the source and image. Yang \etal \cite{Yang2020} detected and cropped out the person's face and used that as an additional image condition within the network for a more fine-grained detailed generation. In addition to human pose, ADGAN \cite{adgan} uses a human parsing map to segment the body parts of the source image to extract their style codes. This allowed them to change the style or texture of clothing region. However, as the shape of the person and clothing is bounded by the segmentation map of the source image, the image edit is limited to only texture transfer. To overcome this issue, PISE \cite{pise} and SPGNET \cite{spgnet} trained a separate network to generate a parsing map of the target pose, which they edited before feeding into the image generator. Allowing the changing of clothing shape  \eg, from short sleeve to long sleeve, but they cannot perform texture transfer simultaneously. While NTED\cite{nted} and CASD\cite{casd} demonstrated transfer of the entire clothing pieces, they do not provide a method to transfer only the texture. DPTN\cite{dptn} uses two paths - source-to-source and source-to-target, while we require only one path for both trainings. Unlike our approach, existing methods can only edit a subset of clothing texture, shape, or appearance (texture and body), but not all of them. \cite{AlBahar2021} uses SMPL body mesh, which is more computationally expensive to process than our method, which uses only 72 parameters. Concurrent to our work, PIDM\cite{persion_dm} shows clothing style interpolation by interpolating the DM's noises, but they could not perform pose interpolation. 

\textbf{Text-Guided Image Generation}. 
There exist text-to-image models since the early days of GANs \cite{stackgan, attngan, dmgan}, to transformer\cite{transformer}-based DALL-E\cite{dalle} and diffusion models\cite{glide,dalle2,ldm, imagen}. However, they do not provide precise control over human pose or fine-grained appearances. KPE\cite{kpe} created the first text-and-pose-guided image generative model that encodes body keypoints into transformer tokens as conditions. Although it can generate accurate poses, as text is a weak condition, it cannot provide fine-grained appearance control and consistency for pose transfer. Using a parsing map and hierarchical autoencoder to encode different body regions, Text2Human\cite{text2human} offered more fine-grained appearance control. Still, it could not specify person and clothing attributes not labeled in the text description, notably the clothing color. HumanDiffusion\cite{human_diffusion} segments and encodes each clothing item with CLIP image encoder into style code and uses a fixed-size database to store the embedding of fashion styles. During sampling, they use either CLIP image or text embedding, but not both, to retrieve the closest embedding from the database. Although this allows them to control the clothing color using text, their method entangles the clothing type, color, and texture pattern into a finite number of combinations. In contrast, our method offers disentanglement and allows users to edit each clothing attribute independently using combination of image and text. The existing generative methods could not consistently generate images for pose transfer or appearance editing. More recently, ControlNet \cite{controlnet} adds pose guidance to DM, but it cannot ensure appearance consistency due to the lack of visual conditioning.
%-------------------------------------------------------------------------
\vspace{-4mm}
\section{Methodology}
\begin{figure*}[!htb]
    \begin{center}
        \includegraphics[width=1.0\linewidth]{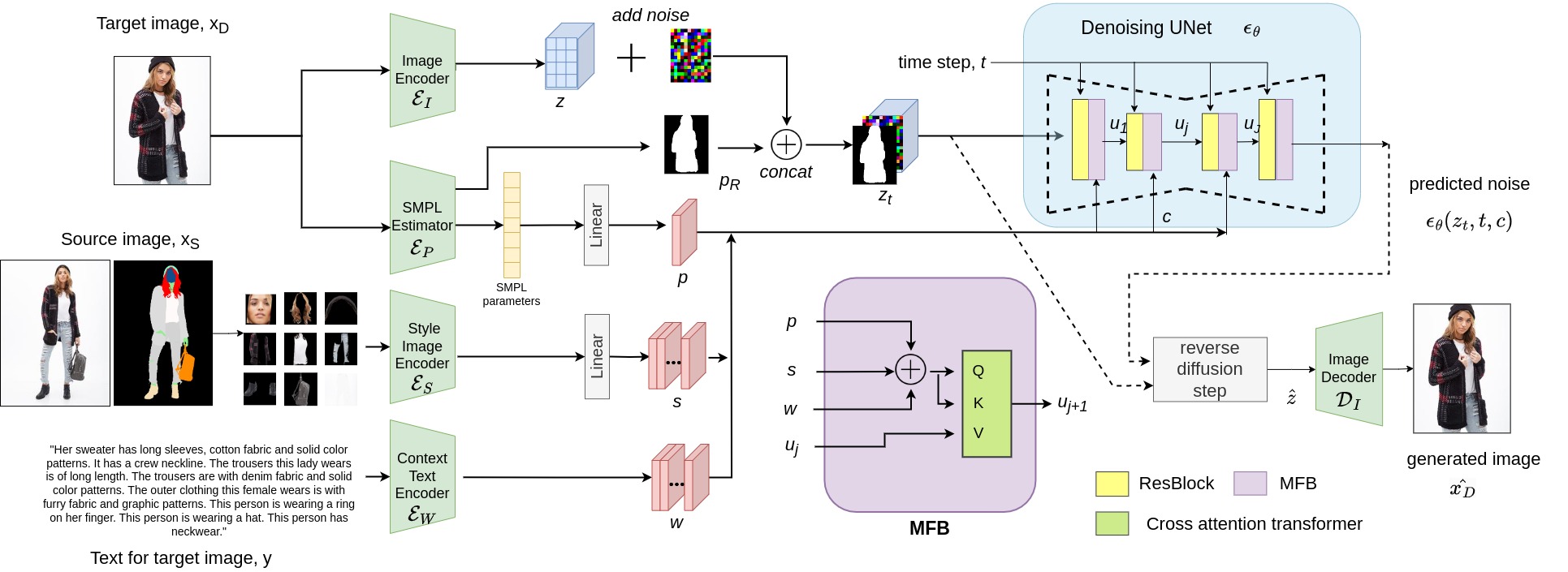}
    \end{center}
    \caption{Overview of our proposed UPGPT architecture. In training, we encode pose, style image, and context text into embeddings that go to the Multimodal Fusing Block (MFB) for fusing. The output of MFB is used as a condition in UNet to predict the noise needed to denoise the image's latent. In sampling, the image encoder decodes the denoised latent $\hat{z}$ into pixel space.}
    \label{fig:model_1}
\end{figure*}

The primary motivation of our proposed method is to fully disentangle a person's image into content and style represented by pose, text, and image features. We can independently edit and mix the different modalities at source to provide fine-grained person image generation and editing. Figure \ref{fig:model_1} illustrates the overall architecture of UPGPT. The first step is to extract the person's information from images and text in the form of features and encode them into conditioning embeddings. The second step is to fuse the embeddings within Multimodal Fusion Block (MFB) to provide conditioning to the UNet of the DM. The figure shows the training pipeline for the pose transfer task with the source-target image pair at the input. However, this can be repurposed for image generation tasks using the same image as the source and target image. Existing person image generation methods \cite{kpe, text2human, human_diffusion} use only individual images in training, while image pairs are necessary for pose transfer methods \cite{patn, Yang2020, gfla, adgan, pise, casd, nted, persion_dm}. Our novel architecture allows us to use individual and paired images to increase the training sample size. The following section describes the proposed method in detail.

\subsection{Multimodal Feature Representation}
Our model uses three modalities: pose, image, and text. We further divide the text into context text and style text. Overall, a person's image is disentangled into content represented by pose and context text; and style as defined by style text and image.

\noindent\textbf{Image Latent.}
We encode the target image $x_D \in \mathbb{R}^{ H \times W \times 3}$ using VAE's \cite{vae} encoder into the latent variables $\mathcal{E_I}(x_D) = z \in \mathbb{R}^{\frac{H}{f}, \frac{W}{f}, d_{V}}$ where $d_{V}$ is the VAE's channel dimension, and $f$ is a downsampling factor in the power of two, and $x_D$ and $z$ are only needed in training. In image sampling, the trained DM generates a new image latent $\hat{z}$,  to be decoded by the VAE decoder into pixel space $\hat{x_D}=\mathcal{D_I}(\hat{z})$. Smaller $f$ \eg, 4 gives higher spatial resolution but quadruples the latent size from $f=8$ and thus increases computational effort considerably. Although large $f$ is more computationally frugal, the resulting $z$ has a smaller spatial dimension, which will store more visual details for the same pixel patch. As a result, a small face in a full-body image can appear blurry after image reconstruction $\mathbb{D}_I(\mathcal{E_I}(x_D))$. 

\noindent\textbf{SMPL Pose}. We use \cite{phosa} as pose estimator $\mathcal{E_P}$ to create an embedding based on the SMPL parameters from the target image $x_D$. The  72 SMPL parameters represent three axis-angle rotations of 24 body joints, ten body shape parameters, and three camera parameters. The camera view of an image is determined by the body's vertical axis rotation parameter and the camera parameters. Each of the three camera parameters in Cartesian coordinate axes determines horizontal translation, vertical translation, and zooming. The SMPL parameters are flattened and projected with a linear layer to $p \in \mathbb{R}^{1 \times d}$ where $d$ is the context text embedding channel dimension. Experiments show that the SMPL's camera parameters are insufficient to ensure the person's correct horizontal position. Therefore, we concatenate a silhouette mask $p_R \in \mathbb{R}^{\frac{H}{f}, \frac{W}{f}} $ at the UNet input to reinforce the pose conditioning, we call it as \textbf{reinforced person mask (RPM)}. RPM only needs to be a coarse mask; this differs from \cite{adgan, pise,spgnet,casd}, which requires a detailed body part segmentation map. We used binary silhouette mask in our main experiments but tried other methods as discussed further in Section \ref{sec:ablation}. 

\noindent\textbf{Style Image}. From a source image $x_S$, we use a segmentation map to segment the person into 9 fine-grained semantic regions \ie, head, hair, headwear, background, top, bottom, outwear, shoes, and bag. Each of the segmented regions is cropped and resized. We call this style image, and we use it as a condition for the person's appearance style. Unlike conventional methods that perform segmentation in run time, we do it in the data preparation stage and store the style regions. This provides image editing flexibility by simply changing the style image files. We do not use source images anymore after obtaining the style images. We treat a person's identity as one of the styles determined by face and hairstyle images. We use a separate face detector to normalize the face - align the face to an upright position. If an occluded face is not detected, we replace it with another normalized face image from the same person if it is available. We encode the style images with a pre-trained CLIP \cite{clip} image encoder $\mathcal{E_S}$ before projecting it with a linear layer into $s \in \mathbb{R}^{N \times d}$  where $N$ is the number of style regions defined for a person.

\begin{figure}[h]
    \vspace{-2mm}
    \begin{center}
        \includegraphics[width=0.8\linewidth]{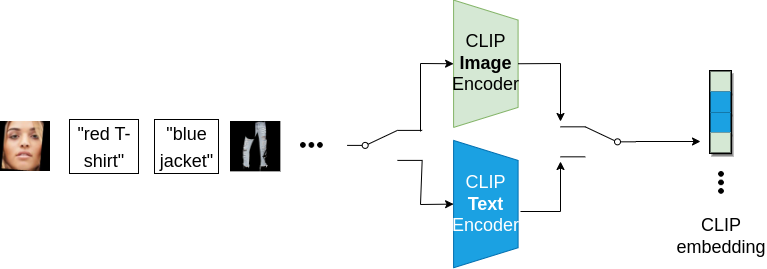}
    \end{center}
    \vspace{-2mm}
    \caption{We can mix-and-match a combination of image (green) and text (blue) embedding in sampling time.}
    \label{fig:style_mix}
\end{figure}

\noindent\textbf{Style Text}. CLIP \cite{clip} trains an image encoder and text encoder jointly on image-text pairs, with a common embedding for both modes aiming to be close to each other in the CLIP embedding space. For example, the CLIP embedding of the text "a red shirt" and an image of a red shirt should be close in terms of Euclidean distance. We use this to create a zero-shot learning method through editing with text. Like us, HumanDiffusion\cite{human_diffusion} uses CLIP image encoding in training, but they can only use either text or image to control image sampling, while we can use either or both modalities. Also, we use two different text conditions - content and style to provide better disentanglement and finer control. Figure \ref{fig:style_mix} shows how we can mix the style images and texts in image sampling. Style text provides a fast and convenient way to control the clothing texture and color, while we can use style images to dictate specific appearances such as face and color shade.  

\noindent\textbf{Content Text.} The content text describes the content of the target image \eg, gender, clothing shapes, and fabrics. We use a pre-trained LLM (large language model) transformer \cite{huggingface_llm} for text encoding. We take the transformer's last layer feature as our content text embedding $\mathcal{E_W}(y)=w \in \mathbb{R}^{l \times d}$ where $l$ is the maximum text token length. 

\subsection{Conditional Diffusion Model}
The DM training process consists of a sequence of time steps $t=1...T$, where Gaussian noise $\epsilon$ is scaled using a noise schedule \cite{diffusion_model} and added to an image latent variable $z$ to produce a noisy version. This concatenates with $p_R$ to produce $z_t$, fed into the input of a denoising UNet $\epsilon_{\theta}$. We propose to condition using our $MFB$ block to concatenate $\oplus$  pose $p$, text $w$ and style embedding $s$ and perform cross-attention with  UNet's ResBlock output at every level $u_j$ where $j$ is layer number.
\vspace{-2mm}
\begin{equation}
    c = p \oplus s \oplus w, Q = \phi_Q(c),K= \phi_k(c),V = \phi_V(u_j)
\end{equation}
where $\phi$ performs $1\times1$ convolution layers for projection into $u_j$'s channel dimension $d_j$ and flatten to 1-dimension.
\vspace{-2mm}
\begin{equation}
    CrossAtten(Q, K, V) = softmax(\frac{QK^T}{\sqrt{d_j}}) V
\end{equation}
\vspace{-2mm}
We train the UNet by using MSE loss on predicted noise $\epsilon_{\theta}(z_t, t, c)$:
\begin{equation}
    \mathcal{L_{MSE}}  := \mathbb{E}_{z,p_R, c,t,\epsilon \sim \mathcal{N}(0,1)} \left[\| \mathcal{W}\odot(\epsilon -\epsilon_{\theta}(z_t, t, c) \|^2_2)\right]
    \label{eq:mse}
\end{equation}
Where $\odot$ is element-wise multiplication and $\mathcal{W} \in \mathbb{R}^{\frac{H}{f}, \frac{W}{f}} $ is loss weight we add to the standard diffusion loss. In addition to the primary loss, many GANs\cite{pise, gfla, casd, nted} use perceptual loss\cite{perceptual_loss}, which extract features from image pixels. However, a single training step in the DM does not generate an image; therefore, we cannot directly use additional losses that require image pixels. Consequently, we use a loss weight $\mathcal{W}$, a 2D tensor with the same dimension as the image latent, to assign different weights to the loss. This helps to regulate the training under challenging regions such as face and hands.

\subsection{Generation, Transfer \& Editing of Images}
Unlike previous pose transfer work, we do not need to use a segmentation map or any reference person image when sampling a new image. We create a new random image latent $z_0$ to begin the sampling process. Progressively in each time step $t$, the image latent is denoised using the reverse diffusion step as described by \cite{diffusion_model} to produce a less noisy image latent $\hat{z_t}=G(z_t, t, c)$. After the $T$ steps, the denoised $\hat{z}$ is decoded by the VAE decoder $\mathcal{D_I}(\hat{z})$ to create an image in pixel space. We use the same pipeline for all the tasks by changing only the conditioning.

To adjust the clothing texture and color, we can either do a texture transfer by using a style image or by replacing the style embedding for that clothing with style text, all without a segmentation mask. Due to the suitable disentanglement property of our method, this changes only the texture and color but not the clothing shape, as demonstrated in the left image in Figure \ref{fig:banner}(c). If we fix the style condition and change only the context text \eg, from "long sleeve" to "short sleeve," it will only change the sleeve length while maintaining the clothing texture. We can modify the content text and style for appearance edit/transfer, which change/copies both the shape and styles. To perform pose transfer, we replace the pose of the source image with one from the target image; this would produce results similar to existing pose transfer methods. On top of that, we use context text from the target image that better describes the desired appearance to generate images with clothing appearance more faithful to the target image. The different configurations are summarized in Table \ref{table:edit}, and some image examples are shown in Figure \ref{fig:banner}.

\begin{table}[]
\begin{center}
\begin{tabular}{l|c|c|c} 
\toprule
\textbf{Task\textbackslash Condition}  & \textbf{Styles} & \textbf{Content} & \textbf{Pose} \\
& &\textbf{Text} & \\
\toprule
\textbf{Generate} & source & source & source \\
%\midrule
\textbf{Texture Edit}  & style image/ & source & source \\
%&style text & & \\
%\midrule
\textbf{Shape Edit}  & source &  target/edit & source \\
%\midrule
\textbf{Appearance Edit}  & style image/  & target/edit & source \\
%                                            & style text  & & \\
%\midrule
\textbf{Pose Transfer}  & source & target & target  \\
\bottomrule
\end{tabular}
\caption{Starting from image generation using information from the source image, the table shows how our method can perform various tasks using different conditioning combinations.}
\label{table:edit}
\end{center}
\end{table}
\vspace{-4mm}
%-------------------------------------------------------------------------
\begin{figure*}[ht]
\centering
    \begin{subfigure}[b]{0.8\textwidth}
    \includegraphics[width=1.0\linewidth]{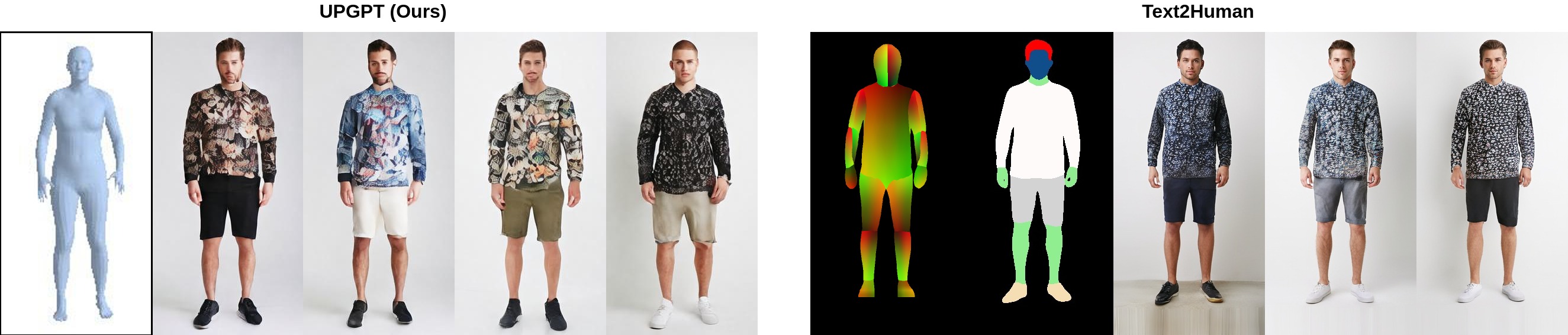}
    \caption{``The gentleman is wearing a long-sleeve shirt with floral patterns and short pants with pure color patterns.''}
    \label{fig:synthesis_man}
    \end{subfigure}    
    \begin{subfigure}[b]{0.8\textwidth}
    \includegraphics[width=1\linewidth]{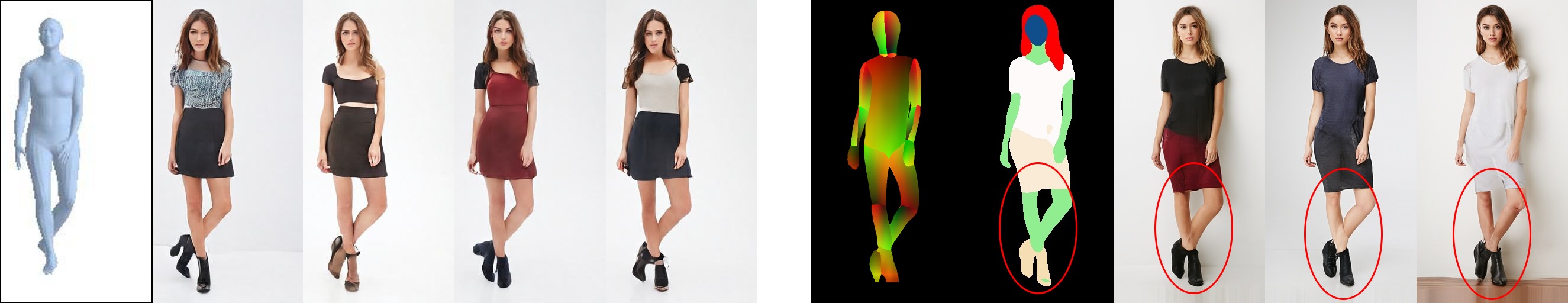}
    \caption{``The woman wears a short-sleeve shirt and short skirt in pure color.''}
    \label{fig:synthesis_leg}    
    \end{subfigure}
    \begin{subfigure}[b]{0.8\textwidth}
    \includegraphics[width=1\linewidth]{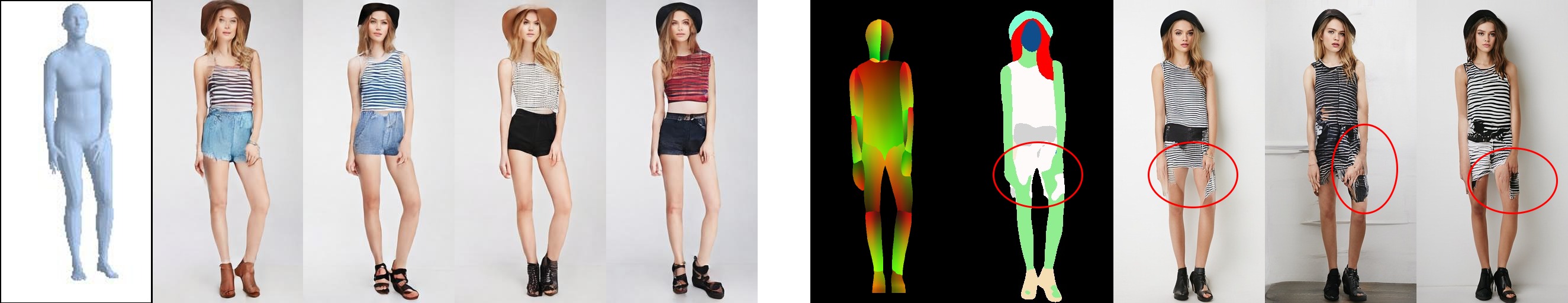}
    \caption{``The lady is wearing a sleeveless shirt, a short pant, and a hat.''}
    \label{fig:ssynthesis_pant}
    \end{subfigure}      
\caption{\textit{(Zoom in to view full $512\times256$)} resolution. (a) We generate a variety of clothing types and texture patterns directly from SMPL pose parameters while Text2Human has additional stage to create parsing map from pose (DensePose\cite{densepose}). (b) Text2Human tend to generate blended crossed legs when the parsing map overlapped. (c) Using vocabulary outside of Text2Human limited dictionary can result in defective parsing map and hence erroneous final image.}
\label{fig:synthesis}
\end{figure*}

\section{Experiments}
We performed experiments on two tasks: (1) text-pose guided image generation and (2) pose transfer. Both use the same model architecture but different image resolutions and subsets of the DeepFashion dataset \cite{deepfashion}.
\\\\
\noindent\textbf{Implementation Details.}
We train our model using AdamW optimizer \cite{adamw} at a learning rate of 5$\times10^{-5}$, batch size of 24, and loss weight, $W$ (Equation \ref{eq:mse}) used is face=8.0, arms=2.0, background=0.5 and 1.0 for others, and a silhouette mask is used for reinforced pose mask. Our model is trained with $T=1000$ noising steps and a linear noise schedule.  
\\
\noindent\textbf{Evaluation Metrics.} 
We use \textit{LPIPS}\cite{lpips} and \textit{SSIM} \cite{ssim} to measure the similarity between the generated image and target image in the pose transfer task. LPIPS uses pre-trained VGG\cite{vgg} to calculate the perceptual similarity, while SSIM measures the similarity by considering the images' luminance, contrast, and structure. For the text-pose guided image generation task, clothing color changes can significantly impact the similarity score, even if it looks realistic. Therefore, instead of comparing individual images, we use Frech\'et Inception Distance (\textit{FID})\cite{fid} to measure the distribution of two groups - ground truth and generated images.

\subsection{Text-Pose Guided Image Generation}
\vspace{-3mm}
\begin{table}[!htb]
\begin{center}
\begin{tabular}{l|c} 
\toprule
\textbf{Method} &  \textbf{FID}$\downarrow$  \\ 
\toprule
 \textdagger HumanDiffusion\cite{human_diffusion} & 30.42    \\
 Text2Human\cite{text2human} & 24.52    \\
\rowcolor{gray!20}\textbf{UPGPT}(Ours)  & \textbf{23.46}  \\
 
\bottomrule
\end{tabular}
\caption{Quantitative result on DeepFashion Multimodal dataset on text-and-pose guided image generation. \textdagger \hspace{1mm} taken from \cite{human_diffusion}.}
\label{table:synthesis}
\end{center}
\end{table}

We use the DeepFashion Multimodal dataset proposed by Text2Human \cite{text2human} in which a segmentation map and text description accompany each image. We train on the resolution $512\times352$. We follow Text2Human's data split and crop the generated images into $512\times256$. The baseline methods \cite{human_diffusion,text2human} cannot control clothing color, which would hugely affect evaluation scores. For a fair comparison, we train our models without clothing style image embedding. 

Table \ref{table:synthesis} shows our method achieving the best FID score against the baselines. Next, we perform some qualitative analysis. HumanDiffusion\cite{human_diffusion} does not provide code to reproduce their results, but their paper shows blurry images with color saturation. Both us and Text2Human can generate high quality images, as shown in Figure \ref{fig:synthesis_man} and Figure B.1 in the appendix, but there are a few shortcomings with the latter. Text2Human cannot generate images directly from the pose, and it must first generate a parsing map from the pose and text. As also observed by \cite{human_diffusion}, we found that they systematically exhibit blended crossed leg when parsing map overlapped (Figure \ref{fig:synthesis_leg}). Parsing maps can also induce gender bias, as detailed in the appendix. Also, Text2Human has limited text capability. Their model was trained on categorical labels and added text-to-category mapping later. Therefore, vocabulary falling outside of their dictionary can generate the wrong parsing map. This is demonstrated in Figure \ref{fig:ssynthesis_pant}. The word \textit{pant} rather than \textit{pant\textbf{s}} was used in the text prompt, and that causes the skin (green) and top clothing (white) to smear into bottom clothing (gray). The following sections will show our superior visual and text prompting capability.

\begin{figure*}[ht]
\begin{center}
\includegraphics[width=0.9\linewidth]{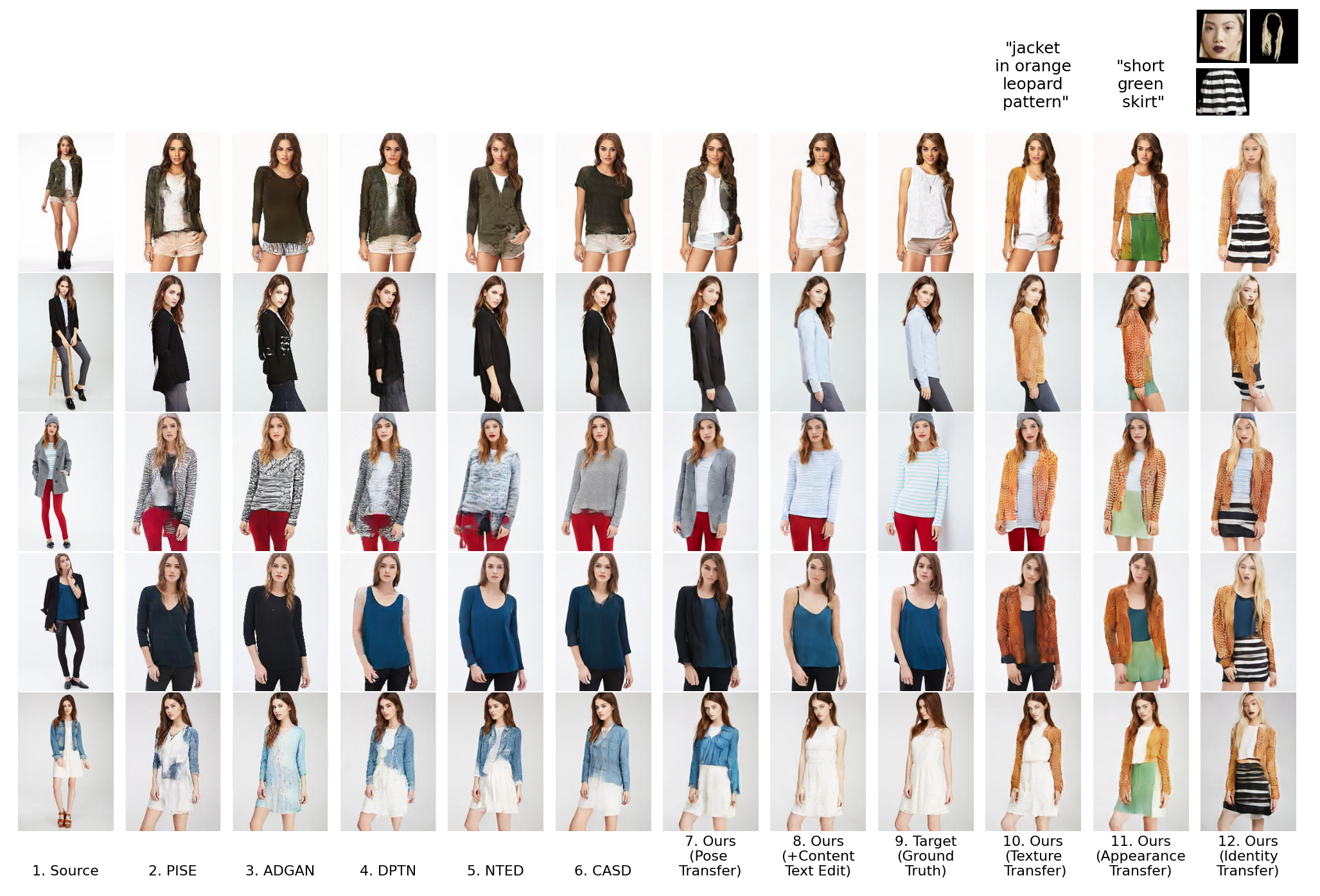}
\caption{\textit{(Zoom in to view)} Pose transfer from (1) source image into the (9) pose target in which the jacket is removed. Reference methods PISE\cite{pise}, ADGAN\cite{adgan}, DPTN\cite{dptn}, NTED\cite{nted}, CASD\cite{casd} blend the top wear and jacket to generate the wrong clothing (2-6), while ours (7) create clear separated jacket from top wear, matching the source image appearance. Conditioning on the content text that correctly describes the target image, we create the final pose transfer result in (8) matching the ground truth (9) appearances. (10) and (11) show we can perform consecutive texture and appearance transfers with texts. In (12), we show how to perform texture and identity transfer using style images while still conditioning on the previous style text edit. }

\label{fig:pose_transfer}
\end{center}
\end{figure*}
\vspace{-4.5mm}
\subsection{Pose Transfer}\label{sec:pose_transfer}
\vspace{-2mm}
We use DeepFashion\cite{deepfashion} In-shop Clothes Retrieval dataset for the pose transfer task. Using the given train-test split of individual images (48675 and 4039, respectively), PATN\cite{patn} proposes a pose transfer dataset of about 102k image pairs for training and 8570 pairs for testing. Given our model architecture's flexibility to support individual and image pairs in training, we combine both as our training dataset. As the Inshop subset does not provide a text description of images, we use the text labels from the Multimodal subset, which cover most of the samples in Inshop. We resize the image to 256$\times$176, maintaining the same aspect ratio. We also combined the fine-grained segmentation map from both subsets. However, a small number, about 5\% of Inshop test image pairs, either have incomplete text or segmentation maps or do not contain humans; we excluded these from the test set. We evaluate our and reference methods using the same reduced test set to obtain the fair quantitative results in Table \ref{table:pt_result}. 

\begin{table}[]
\begin{center}
\begin{tabular}{l r c c} 
\toprule
\textbf{Method} &  \textbf{FID}$\downarrow$ & \textbf{LPIPS}$\downarrow$ & \textbf{SSIM}$\uparrow$   \\ 
\toprule
 ADGAN\cite{adgan} & 20.025 & 0.2289 & 0.6856 \\
 PISE\cite{pise} & 17.799 & 0.2273 & 0.6781 \\
 DPTN\cite{dptn} & 16.686 & 0.2192 & 0.6958 \\
 CASD\cite{casd}  & 10.439 & \textbf{0.1777} & \textbf{0.7131} \\
\rowcolor{gray!20}  UPGPT(ours) & 9.427 & 0.1886 & 0.6970 \\
 NTED\cite{nted}  & \textbf{8.813} & 0.1814 & 0.7011   \\
 \rowcolor{gray!20} \textdagger UPGPT(ours) & 7.876 & 0.1766 & 0.7276 \\
\bottomrule
\end{tabular}
\caption{Quantitative results on pose transfer task.  \textdagger  \hspace{1mm} compare the generated images against images reconstructed by VAE.}
\label{table:pt_result}
\end{center}
\end{table}

Although our method is not designed explicitly for the pose transfer task alone, we near state-of-the-art results; we found that small faces in our generated images can appear blurry due to the inadequacy of VAE in capturing rich details in small faces. In other words, an image $x$ reconstructed $\mathcal{D_I}(\mathcal{E_I}(x))$ by VAE can have a blurry face even if our model produces a perfect image latent. To confirm this, we compare our generated images against the images reconstructed from the ground truth images rather than the ground truth images, and the scores improve significantly to top the performance table. 

Apart from that, our method produces realistically looking images and excels in utilizing all modalities when information in the source image is incomplete or incorrect. This is best demonstrated in the pose transfer task in Figure \ref{fig:pose_transfer}, where the jacket in the source image (1) is removed from the target image (9). Even assuming the person still has their jacket on, existing methods (2-6) often fail to distinguish between the jacket and the top wear, blending the style and texture to create incorrect clothing. In contrast, our results (7) show clear distinguishment, resembling the source image appearance. UPGPT blocks out the jacket in the image generation process by conditioning on the context text of the target image, creating our final pose transfer result (8) that resembles the ground truth target image (9). 

\subsection{Flexible Image Editing}
Columns (10-12) in Figure \ref{fig:pose_transfer} demonstrate the flexibility of our fine-grained control method. From (7), we replace the jacket style image with the style text ``jacket in orange leopard pattern'' to perform texture transfer (10). Our style text has good zero-shot capability, and we can use words like zebra, pandas, and oceanic instead of color. Then, we change the context and style texts in (11) to replace bottom wear with a short green skirt, changing the texture and clothing type \ie appearance edit. Please note that the jacket from (10) remains in (11), showing that our approach allows for consecutive editing. This is a significant improvement from existing methods \cite{casd, nted, persion_dm} that have demonstrated only to transfer appearance from a single image reference. In contrast, we can mix different modalities from different sources to perform flexible and fine-grained control across clothing type, texture, or both. Although it is convenient to use text to change clothing types and colors, some things are difficult to describe in words \eg specific clothing patterns or the face of a particular person. Therefore, our methods also support using styles images for editing and identity transfer, as shown in (12). The pipeline of going from (1) to (12) demonstrates we can mix and match different modalities - pose, style, content text, and style images to achieve excellent fine mix-and-match in generating and editing person images. 

\subsection{Pose and Camera View Interpolation}
\vspace{-4mm}
\begin{figure}[!ht]
\centering
    \includegraphics[width=1\linewidth]{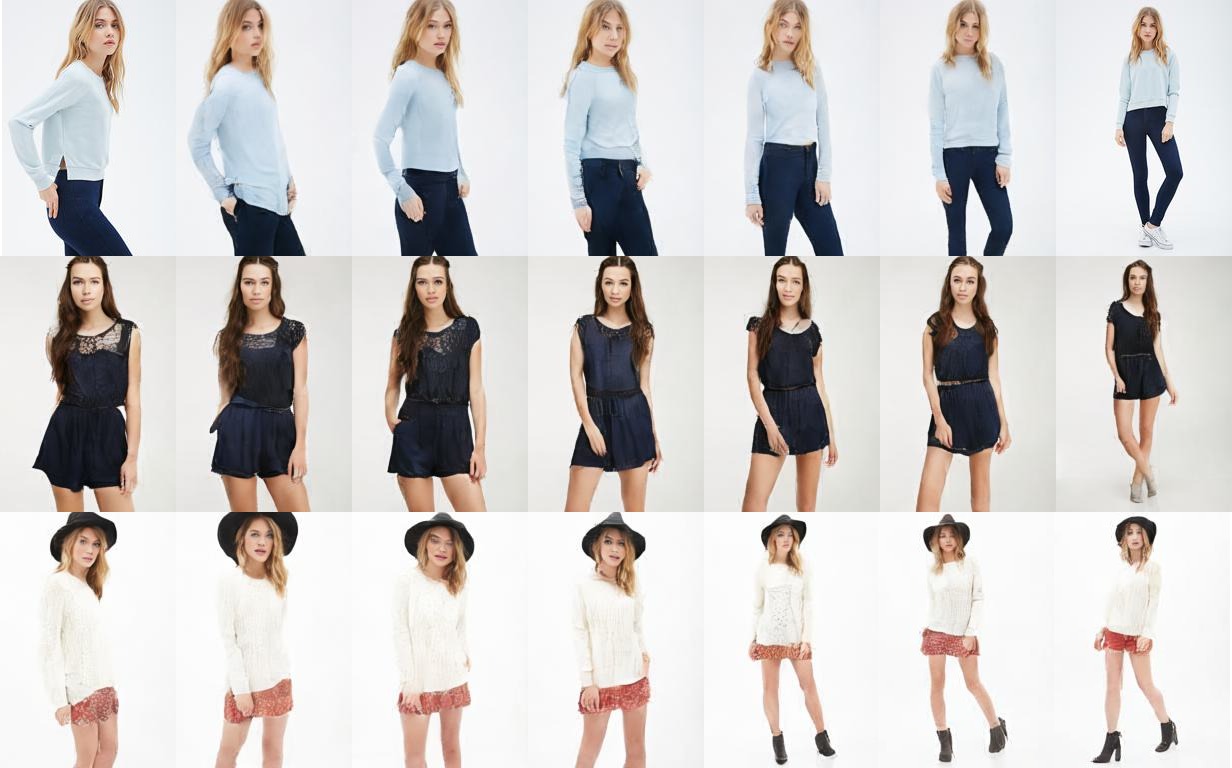}
\caption{Complex hand and camera movement achieved using linear pose interpolation.}
\label{fig:interpolation}
\end{figure}
We demonstrate  our approach's superior pose capability and disentanglement with simultaneous pose and camera view interpolation as shown in Figure \ref{fig:interpolation}. The pose interpolation can be performed by linear interpolating SMPL parameters between two poses. To our best knowledge, this is the first demonstration of pose interpolation within the human image generation literature. 

\subsection{Ablation Study}\label{sec:ablation}
We performed experiments to explore the importance of the reinforced pose masks (RPM) and evaluated their performances as shown in Table \ref{table:ablation}. Qualitatively, without any form of RPM, the person in generated images looks visually similar to other masks apart from the occasional horizontal offset. We explore two methods: a bounding box and a mask derived from the SMPL render. Including a bounding box as a mask improves the scores compared to not having one. A further approach is to use the silhouette mask created from SMPL rendering as the segmentation input. However, the derived mask is less accurate, and the result is slightly worse than using a silhouette mask estimated from 2D images. 

\begin{table}[h]
\centering
\begin{tabular}{l|r|c|c} 
\toprule
\textit{\textbf{Reinforced Pose }} &  \textbf{FID}$\downarrow$ & \textbf{LPIPS}$\downarrow$ & \textbf{SSIM}$\uparrow$   \\ 
\textit{\textbf{ Mask (RPM})} & & & \\
\toprule
  w/o RPM & 10.176 & 0.2670 & 0.6146 \\
w bounding box & 10.100 & 0.2447 & 0.6254 \\
 w SMPL rendering & \textbf{9.245} & 0.2149 & 0.6604   \\
  \textbf{w silhouette mask} & 9.427 & \textbf{0.1886} & \textbf{0.6970} \\
\bottomrule
\end{tabular}

\caption{Quantitative results of ablation on reinforced pose mask.}
\label{table:ablation}

\end{table}

\section{Limitations}
Apart from the blurry small faces discussed in Section \ref{sec:pose_transfer}, one of our method's limitations is that clothing textures sometimes do not match the style image \eg, the stripes can have different thicknesses. This is due to the limitation of the CLIP image encoder, which does not necessarily capture fine-grained spatial detail but focuses on the overall color response. 
%-------------------------------------------------------------------------

%--------------------------------------------------------------------a -----
\section{Conclusion}
In this paper, we proposed UPGPT, the first universal method to perform unified person image generation, editing, and pose transfer tasks. Unlike existing methods that require masks for editing, our mask-less approach provides a convenient way of fine-grained person image editing using a combination of modalities. We achieved competitive pose transfer results in comparison to the state-of-the-art methods. Also, we overcame the inadequacy of SMPL pose estimation to incorporate it into our model to improve pose disentanglement and demonstrate the first simultaneous pose and camera view interpolation in pose-guided image generation literature.

\clearpage
{\small
\bibliographystyle{ieee_fullname}
\bibliography{upgpt}
}

\clearpage
\renewcommand\thesection{\Alph{section}}
\renewcommand\thefigure{\thesection.\arabic{figure}}
\setcounter{section}{0} 
\setcounter{figure}{0} 
\onecolumn
\begin{center}
    {\huge Appendix}
\end{center}
This appendix consists of two sections. Section \ref{sec:B} contains more $256\times176$ examples of our results from Section 4.2 for pose transfer and image edit. Section \ref{sec:A} provides further quantitative analysis on baseline method Text2human using the method detailed in Section 4.1 for image generation task at higher resolution $512\times256$. 

\section{Image Editing}\label{sec:B}
\begin{figure}[H]
    %\begin{center}
        \begin{subfigure}[]{0.5\columnwidth}
        \centering
        \includegraphics[width=0.9\linewidth]{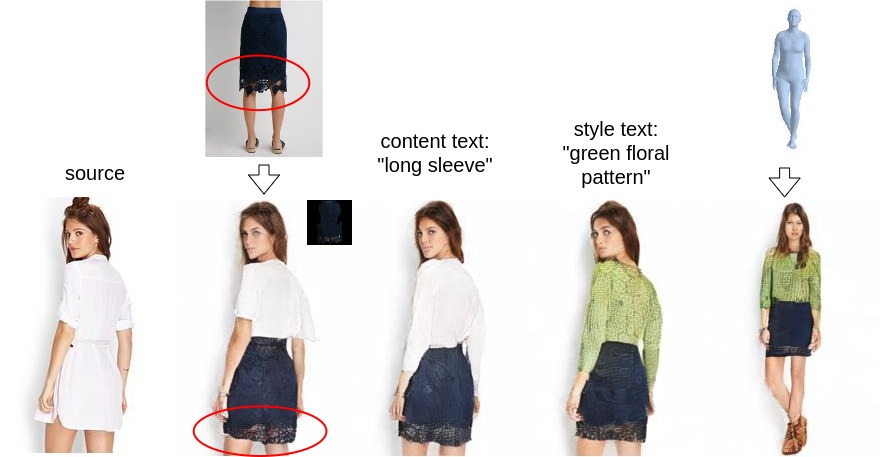}
        \caption{Our method can transfer delicate fashion patterns and pose.}
        \label{fig:appendix.edit3}
        \end{subfigure}    
        \begin{subfigure}[]{0.5\columnwidth}
        \centering
        \includegraphics[width=0.9\linewidth]{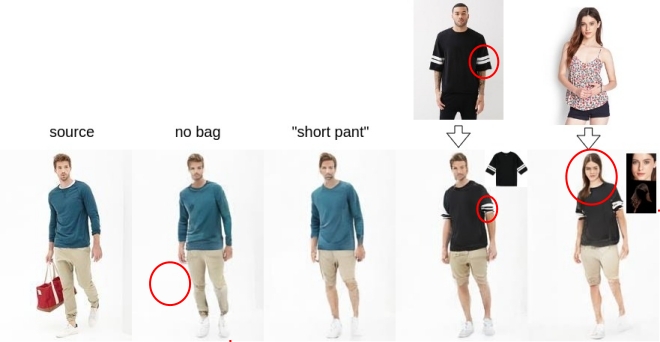}
        \caption{Remove the bag, edit length of pants, transfer clothing pattern and identity \\(face and hair).}
        \label{fig:appendix.edit1}
        \end{subfigure}   
        \begin{subfigure}[]{0.5\columnwidth}
        \centering
        \includegraphics[width=0.9\linewidth]{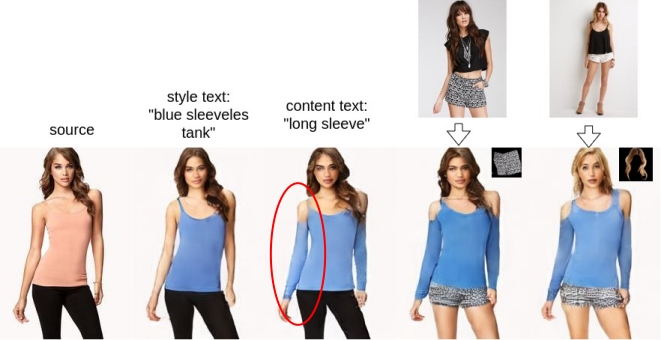}
        \caption{We create a new clothing style by mixing ``sleeveless tank'' in style text with ``long sleeve'' in context text. We can also provide fine-grained transfer of only the hair.}
        \label{fig:appendix.edit2}
        \end{subfigure}          
        \begin{subfigure}[]{0.5\columnwidth}
        \centering
        \includegraphics[width=0.9\linewidth]{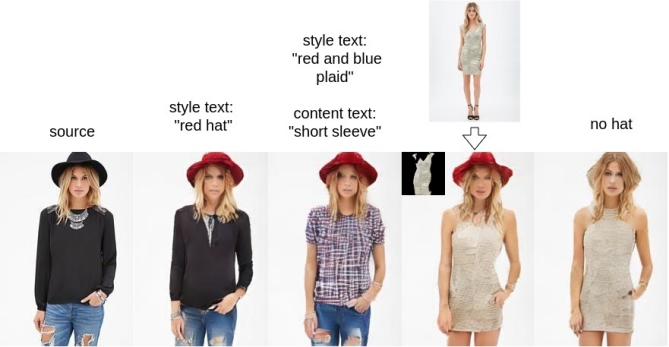}
        \caption{Replacing two garment pieces (shirt and pants) with a single dress.}
        \label{fig:appendix.edit4}
        \end{subfigure}         
                
    %\end{center}
\vspace{1mm}
\caption{Starting from the source image in the left, we perform step-by-step consecutive image editing from multiple multimodal sources.}
\end{figure}

This section provides more image editing examples using a model trained on the low resolution 256$\times$176 images. Our method allows for simultaneous and consecutive image editing using multimodality from multiple sources. Some baseline pose transfer models can perform only a single appearance transfer (clothing texture, shape, or face) from a single reference image, but we could do much more. In Figure \ref{fig:appendix.edit3}, we demonstrate the capability of our method to transfer a delicate clothing style, followed by text edits and pose transfer. We can also remove objects (bag in Figure \ref{fig:appendix.edit1}, hat in Figure \ref{fig:appendix.edit4}). Figure \ref{fig:appendix.edit2} shows how we can create a new  clothing type not from the dataset by mixing "sleeveless tank" in the style text and "long sleeve" in the content text.  Baseline methods are limited to fashion transfer of the same type \ie, top wear to top wear. Still, we can do any combination of fashion transfer, such as replacing a shirt and pants with a dress, as shown in Figure \ref{fig:appendix.edit4}.

\newpage
\section{Text-Pose Guided Generation}\label{sec:A}
\setcounter{figure}{0} 
Overall, Text2Human and our method, UPGPT, can generate high quality images; we display examples of both results and ground truth in Figure \ref{fig:appendix.t2h_1}. Some of Text2Human's images may appear smaller because of the padding they added to the dataset, while we use unmodified DeepFashion Multimodal images. However, Text2Human has two major limitations that can affect the overall visual perception - (1) systematic error in crossed legs and (2) poor gender and pose disentanglement.
\begin{figure}[h]
    \begin{center}
        \includegraphics[width=1.0\linewidth]{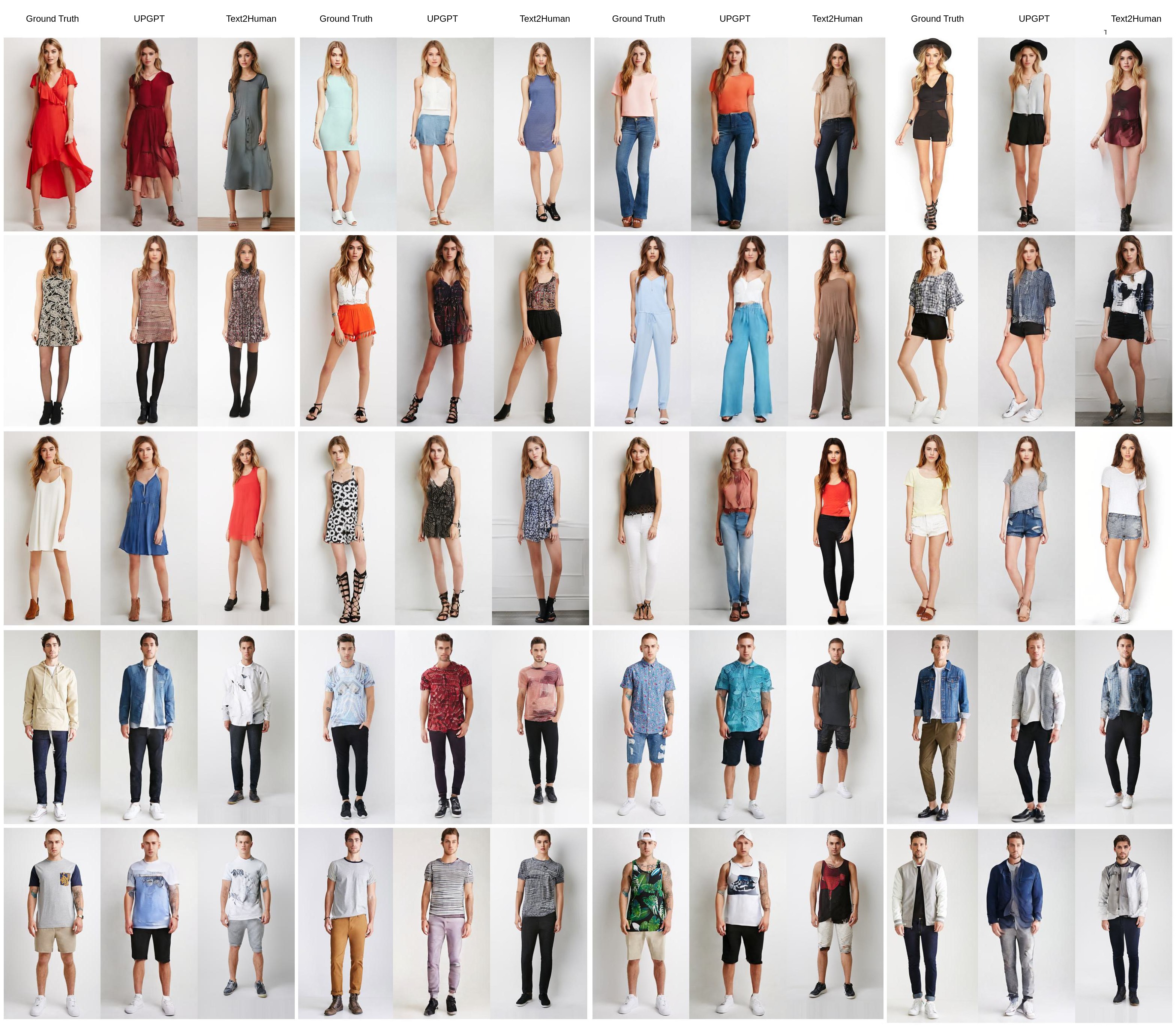}
    \end{center}
\caption{(Zoom in to view full resolution) Both UPGPT(our method) and Text2Human can generate high quality images.} 
\label{fig:appendix.t2h_1}
\end{figure}
\newpage
\subsection{Blended Crossed Legs}
\begin{figure}[h]
    \begin{center}
        \includegraphics[width=0.8\linewidth]{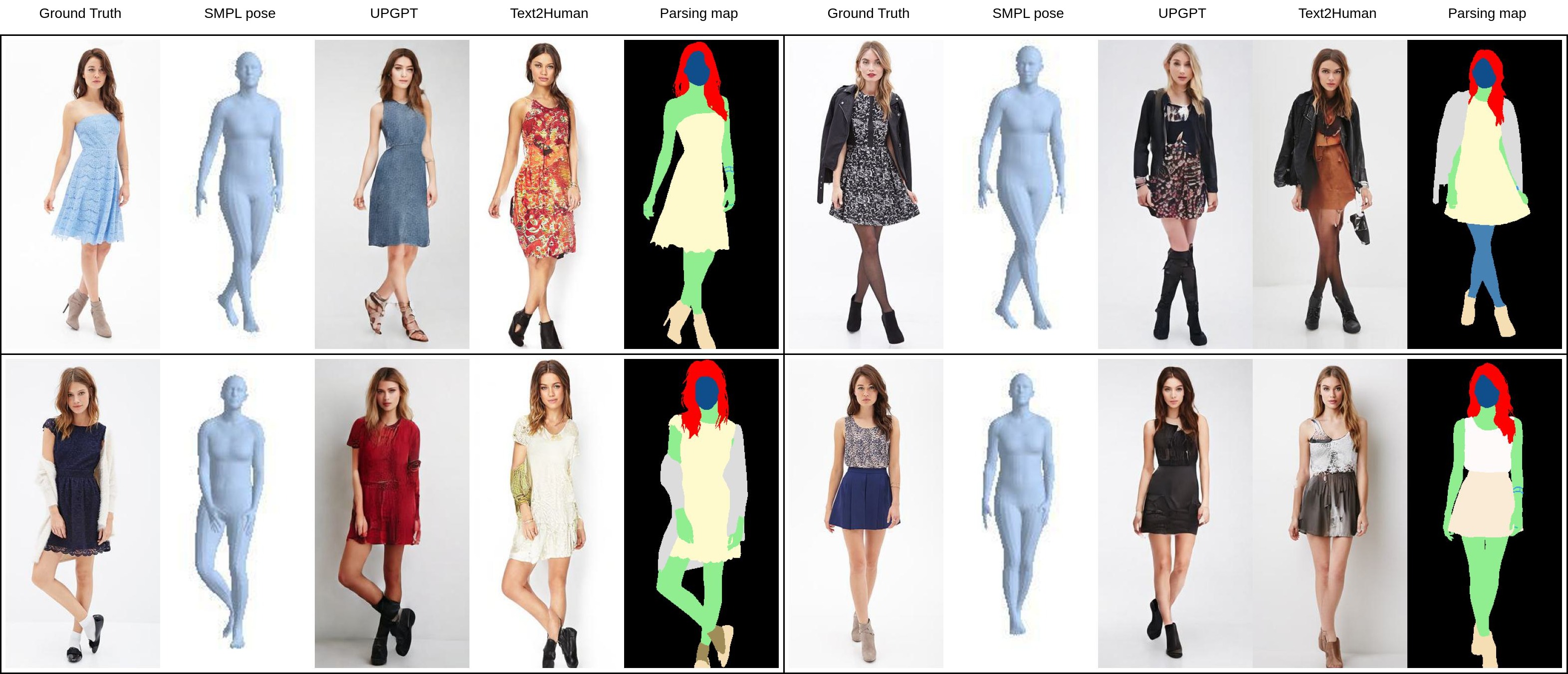}
    \end{center}
\caption{\textit{(Zoom in to view)} Text2Human often generates erroneous crossed legs from parsing map. Our method avoids this problem by using the SMPL model as pose conditioning.}
\label{fig:appendix.leg}
\end{figure}

Figure \ref{fig:appendix.leg} shows systematic error in the legs when crossed and blended in the parsing map and results in the same in the generated images. We avoid this problem by using the SMPL model as pose guidance which contains 3D body pose information. 

\subsection{Poor Gender and Pose disentanglement}
\begin{figure}[H]
    \begin{center}
        \begin{subfigure}[]{0.8\columnwidth}
        \centering
        \includegraphics[width=0.85\linewidth]{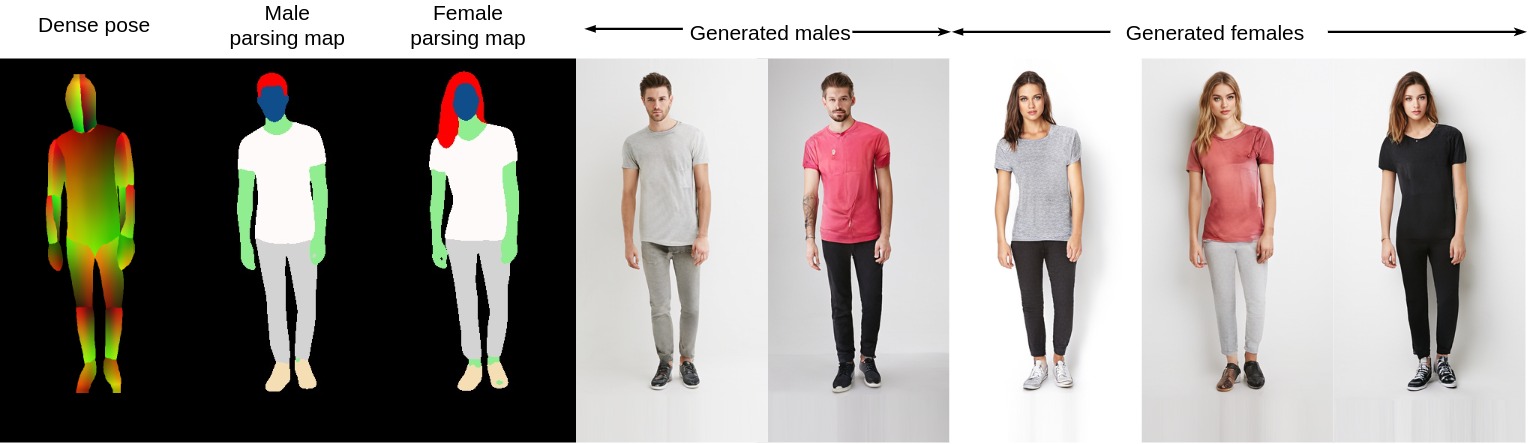}
        \caption{Text2Human. The female appearances have very little difference to males apart from the head. The generated female appear to have broader shoulder than images in dataset.}
        \label{fig:appendix.gender2a}
        \end{subfigure}
        \begin{subfigure}[]{0.8\columnwidth}
        \centering
        \includegraphics[width=0.85\linewidth]{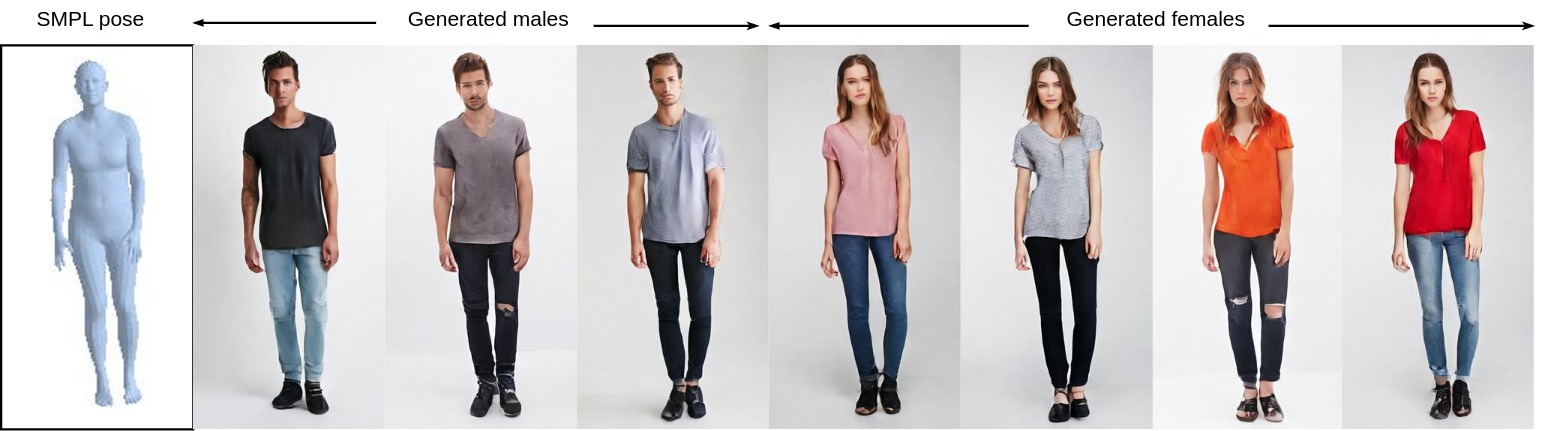}
        \caption{UPGPT. People generated from the same pose look more natural for their genders.}
        \label{fig:appendix.gender2b}
        \end{subfigure}        
    \end{center}
\caption{Our method provides better disentanglement between pose and gender.}
\label{fig:appendix.gender2}
\end{figure}

\begin{figure}[H]
    \begin{center}
        \includegraphics[width=0.8\linewidth]{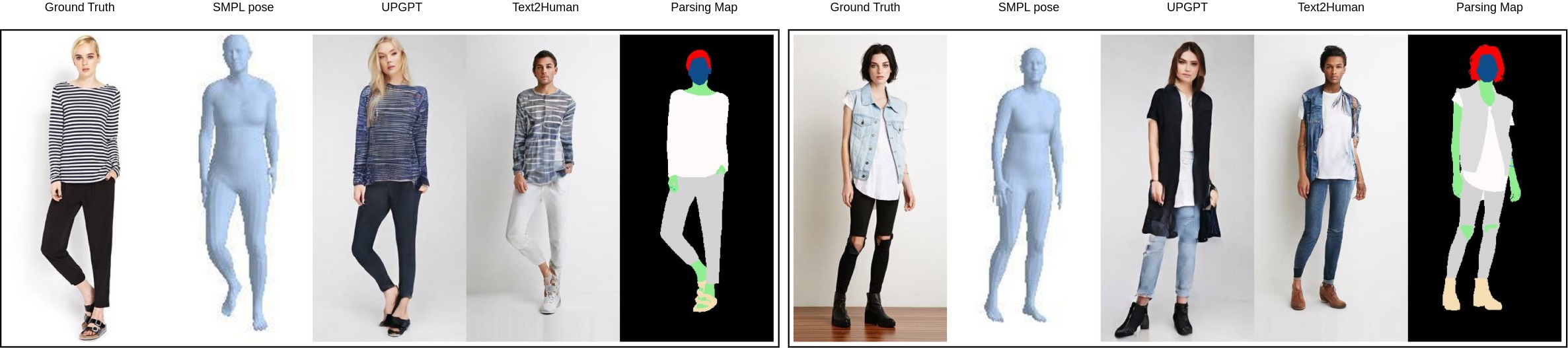}
    \end{center}
\caption{Text2Human tends to generate male faces from parsing maps with short hair.}
\label{fig:appendix.gender1}
\end{figure}

In Text2Human, the body appearance ties closely to the parsing map. Figure \ref{fig:appendix.gender2a} shows that Text2Human generates two parsing maps - male and female from the pose. There is very little difference between them apart from the hair length. Due to incompatible body proportion, Text2Human females' overall appearance (Figure \ref{fig:appendix.gender2a}) have subtle unnaturalness compared to ours in \ref{fig:appendix.gender2b}. Although we use only the female SMPL model to train our model, our model can generalize the genders well yet provide good disentanglement between gender and pose. The gender bias in Text2Human can be further shown in Figure \ref{fig:appendix.gender1} where short haired parsing maps often result in a male face, which doesn't occur with our approach.

\end{document}